\documentclass[10pt,twocolumn,letterpaper]{article}

\usepackage{iccv}
\usepackage{times}
\usepackage{epsfig}
\usepackage{graphicx}
\usepackage{amsmath}
\usepackage{amssymb}

\usepackage[inline]{enumitem}
\usepackage{multirow}
\usepackage{graphicx}
\usepackage{url}

\usepackage[pagebackref=true,breaklinks=true,letterpaper=true,colorlinks,bookmarks=false]{hyperref}

\iccvfinalcopy % *** Uncomment this line for the final submission

\def\iccvPaperID{23} % *** Enter the ICCV Paper ID here

\ificcvfinal\fi
\begin{document}

%%%%%%%%% TITLE
\title{Estimated Depth Map Helps Image Classification}

\author{Yihui He\thanks{This work was done as course project of Advanced Data Mining in UCSB}\\
Xi'an Jiaotong University\\
Xi'an, China\\
{\tt\small heyihui@stu.xjtu.edu.cn}
%\and
}

\maketitle
%\thispagestyle{empty}

%%%%%%%%% ABSTRACT
\begin{abstract}
	We consider image classification with estimated depth. 
	This problem falls into the domain of transfer learning, 
	since we are using a model trained on a set of depth images to generate depth maps (additional features) for use in another classification problem using another disjoint set of images. 
	It's challenging as no direct depth information is provided. 
	Though depth estimation has been well studied \cite{liu2015deep}, 
	none have attempted to aid image classification with estimated depth.
	Therefore, we present a way of transferring domain knowledge on depth estimation to a separate image classification task over a disjoint set of train, and test data. 
	%However, whether the transferred knowledge could enhance classification performance remains a question.
	We build a RGBD dataset based on RGB dataset and do image classification on it.
	Then evaluation the performance of neural networks on the RGBD dataset compared to the RGB dataset.
	From our experiments, the benefit is significant with shallow and deep networks.
	%Unexpectedly, it does harm to performance as network goes deeper.
	It improves ResNet-20 by \textbf{0.55\%} and ResNet-56 by \textbf{0.53\%}.% however drops ResNet-110 0.61\%.
	%To our knowledge, we are the first to bridge gap between image classification and depth estimation.
	Our code and dataset are available publicly.\footnote{\href{https://github.com/yihui-he/Estimated-Depth-Map-Helps-Image-Classification}{github.com/yihui-he/Estimated-Depth-Map-Helps-Image-Classification}}
\end{abstract}

\section{Introduction}
Estimating depths from a single monocular image depicting general scenes is a fundamental problem in computer vision, 
which has widespread applications in scene-understanding, 3D modeling, robotics, and other challenging problems.
It is a notorious example of an ill-posed problem, 
as one captured image may correspond to numerous real world scenes \cite{eigen2014depth}. 
It remains a challenging task for computer vision algorithms as no reliable cues can be exploited,
such as temporal information, stereo correspondences.
Previous research involving depth maps usually involve geometric \cite{hedau2010thinking,gupta2010estimating,gupta2010blocks}, convolutional \cite{liu2015deep} and semantic \cite{ladicky2014pulling} techniques.
Nevertheless, none of these works tried to perform image classification using depth maps as training data.
Different from previous efforts, 
we propose to utilize \textit{estimated depth maps} in a image classification task.
While extensively studied in semantic labeling and accuracy improvement,
depth map regression has been less explored in its application to classification problems. 
Intuitively, one can imagine that a neural-network that is deep enough would generate it's own depth-map (or at least simulate depth-map-like features).

Recently, the efficacy and power of the deep convolutional neural network (CNN) has been made accessible \cite{he2015deep,jia2014caffe}. 
With a CNN, we are able to perform depth estimation on a single image \cite{liu2015deep}.
However, most classification tasks still perform on RGB images.
With only RGB images, CNN features have been setting new records for a wide variety of vision applications \cite{he2015deep,razavian2014cnn,girshick2015fast,fasterrcnn,dai2016instance}.
Despite all the successes in depth estimation and image classification,
the deep CNN has been not yet been used for learning on RGBD images, since RGBD datasets are not as widely-used as RGB datasets.

\begin{figure}
	\centering
	\includegraphics[width=.8\linewidth]{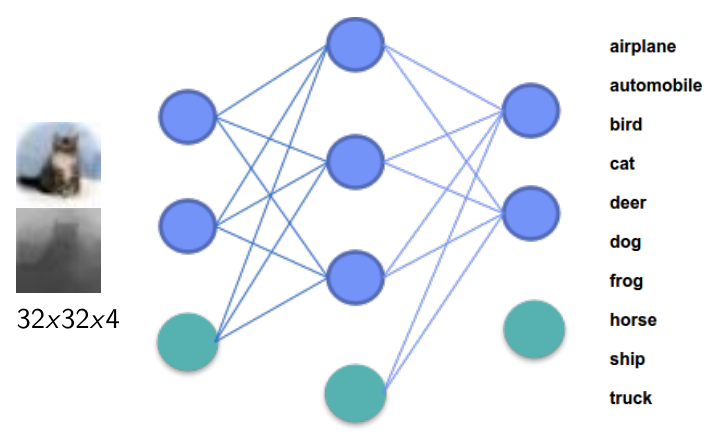}
	\caption{Learning on RGBD dataset: First we generate the depth map of an image with a depth estimation network (On the Left). 
		Second, we perform image classification on the depth map and the image.}
	\label{fig:tarch}
\end{figure}

We propose to build a RGBD dataset based on RGB dataset and do image classification on it, illustrated in Figure~\ref{fig:tarch}.
Then evaluation the performance of neural networks on the RGBD dataset compared to the RGB dataset.
To our knowledge, we are the first to bridge gap between \textit{estimated depth} and image classification.
From our experiments, the benefit is significant on both shallow and deep networks.
%Unexpectedly, it does harm to performance as network goes deeper.
It improves ResNet-20 \textbf{0.55\%}. %, however drops ResNet-110 0.61\%.

To sum up, we highlight the main contributions of this work as follows:
\begin{itemize}
	%\vspace{-.12cm} 
	%\item We implemented a deep convolutional neural field in order to solve the depth estimation problem and obtained similar results to the implementation used in the paper.
	%\vspace{-.12cm} 
	\item We created a RGBD image dataset for CIFAR-10.
	%\vspace{-.12cm} 
	\item We illustrate that depth channel has a better feature representation than R,G,B channels,
	and show that training on RGBD images could improve performance.
	\item We define a new metric for ill-posed depth prediction problem.
	%\vspace{-.12cm} 
\end{itemize}

%-------------------------------------------------------------------------
\section{Related Work}
CNN have been applied with great success for object classification \cite{krizhevsky2009learning,simonyan2014very,szegedy2015going,he2015deep,carreira1998xception} and detection \cite{girshick2015fast,fasterrcnn,dai2016instance}. 
CNN have recently been applied to a variety of other tasks, like depth estimation. 
Depth estimation from single image is well addressed by Liu \etal\cite{liu2015deep} and Eigen \etal\cite{eigen2015predicting}. 
They both agree that depth estimation is an ill-posed problem, 
since there's no real ground truth depth map.
We define transfer learning accuracy metric for depth estimation model (Section~\ref{sec:dataset}). 
It becomes easier to compare performance of different depth estimation model.

Estimated depth map \cite{liu2015deep} has been successfully applied to some other problems. 
Based on depth information, performance improved on semantic labeling \cite{eigen2015predicting}. 
However, depth maps have not been combined with an image classification task. 
To our knowledge, we are the first to bridge gap between depth estimation and image classification.

There are already many successful transfer learning results in Computer Vision.
A popular one is transfer ImageNet \cite{deng2009imagenet} Classification Network like VGG-16 \cite{simonyan2014very} to object detection \cite{fasterrcnn}.
Another example also in object detection is contextualized networks \cite{li2017attentive}, usually through multi scale context.

\section{Approach}
Recent depth image research works mainly focus on depth-estimation \cite{liu2015deep} 
and segmentation with depth image \cite{eigen2015predicting}.
And we've witnessed significant improvement on depth estimation quality in these years. 
However, most image classification tasks nowadays are still performed on RGB images.
So we want to transfer depth knowledge learned by depth estimation model into our image-classification model.
In this section, we first built a RGBD dataset for CIFAR-10 \cite{krizhevsky2009learning}, 
based on a trained deep convolutional neural field model \cite{liu2015deep}. 
To investigate the effect of the depth channel on image classification task, 
we design two experiments (one with a simple feedforward NN and one with a CNN) 
Finally, we propose a new metric for depth estimation performance measurement.

\subsection{Build RGBD Dataset} \label{sec:dataset}
\begin{figure}
	\centering
	\includegraphics[width=.8\linewidth]{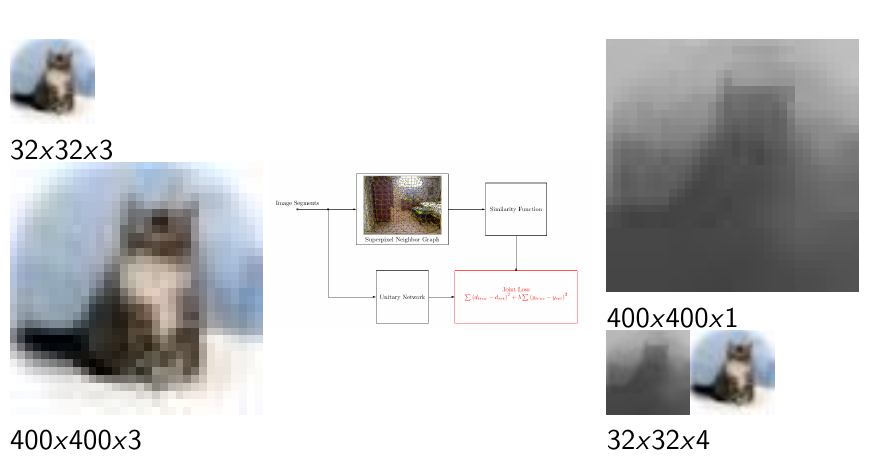}
	\caption{Transer Learning: Build RGBD CIFAR-10 dataset}
	\label{fig:dataset}
\end{figure}

\begin{figure}
	\centering
	\includegraphics[width=.8\linewidth]{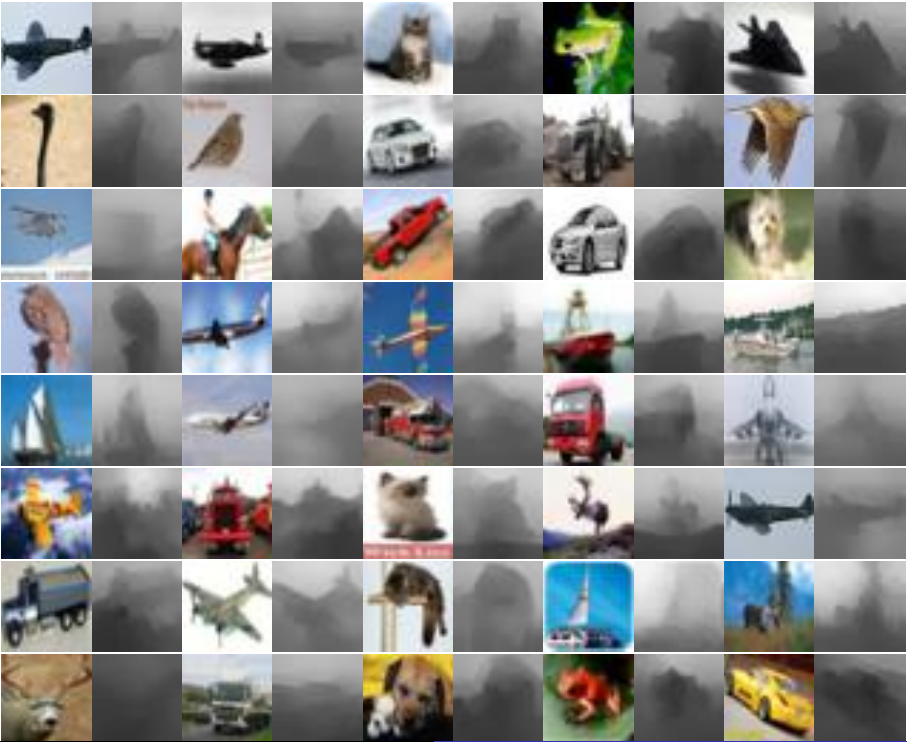}
	\caption{Depth map estimated by deep convolutional neural field}
	\label{fig:tiny}
\end{figure}

Since the Deep Convolutional Neural Field model accepts images \cite{deng2009imagenet} that are 
much larger that CIFAR-10 tiny images ($32\times32$), we build RGBD dataset as follow:
\begin{enumerate}
	\item resize CIFAR-10 tiny image ($32\times32\times3$) to normal size ($400\times400\times3$) in order to feed in CNF.
	\item perform depth estimation on the normal size image.
	\item downscale the output image (depth image, $400\times400\times1$) back to tiny image ($32\times32\times1$).
	\item combine RGB and D channels together as our RGBD image ($32\times32\times4$).
\end{enumerate}
Figure~\ref{fig:dataset} shows the transfer learning procedure.
Figure~\ref{fig:tiny} shows some depth maps.
Since there is no ground-truth depth image for CIFAR-10 dataset, we can't directly 
measure the quality of our depth estimation attempts for these tiny images. 
However, we can infer this indirectly.% in two ways.
%First we can look at these depth images
%and make sure that most of them is reasonable.
We can use the accuracy results of our two experiments as a new metric to quantify depth map quality.

\subsection{Classification Task on RGBD Dataset}
In order to make it easier to show effect of depth channel,
we employ a simple two layer neural network for classification task.
The architecture for learning on the RGBD dataset is shown in Figure~\ref{fig:tarch}. 

The number of neurons in the input layer depends on input.
If input is a single channel (R, G, B, D), we have $32\times32$ neurons.
The amount of hidden neurons is not determined. We perform fine tuning for each situation. 
The number of output neurons is always number of classes (10 classes for CIFAR-10). 
%Details of our architecture is shown in Table~\ref{tab:details}. 
%Hyper-parameters are not discussed here but will be fine tuned.
%\begin{table}
%	\begin{center}
%		\begin{tabular}{|l|l|l|l|}
%			\hline
%			regularization&Activation&Update&batch\\
%			\hline
%			Dropout&ReLU&Momentum&128\\
%			\hline
%		\end{tabular}
%	\end{center}
%	\caption{Architecture details for classification task on our new RGBD dataset}
%	\label{tab:details}
%\end{table}

\section{Experiment}
We measure depth map quality in two ways. 
First, we evaluate neural network performance on R, G, B, D channel as input respectively.
Second, we train neural network on RGB, RGBD respectively and compare the performance.
Our depth estimation is based on Tensorflow \cite{tensorflow2015-whitepaper}.
And our neural network training is based on Caffe \cite{jia2014caffe}.

\subsection{R vs G vs B vs D}
\begin{figure}
	\centering
	\includegraphics[width=.8\linewidth]{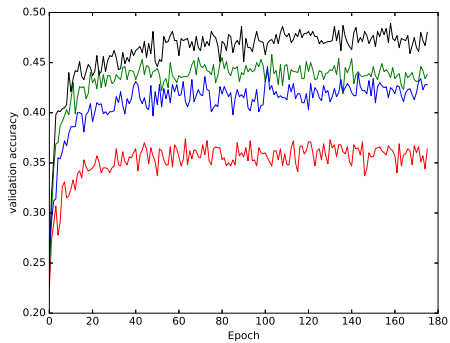}
	\caption{R vs G vs B vs D, testing time}
	\label{fig:channeltest}
\end{figure}

We perform fine tuning on each channel. 
So that their performances are approximately optimal.
%Figure~\ref{fig:channeltrain} shows training accuracy comparison through time. 
Figure~\ref{fig:channeltest} shows validation accuracy comparison through time. 
%\begin{figure}
%	\centering
%	\includegraphics[width=\linewidth]{../presentation/train.eps}
%	\caption{R vs G vs B vs D, training time}
%	\label{fig:channeltrain}
%\end{figure}

You can see that, at testing time, the depth channel outperforms R, G, B channels under the same architecture.
This implies that, depth channel has a better feature representation than R, G, B channels.
Training on RGBD dataset would result in better performance for shallow networks.
%We verify whether this is true in the next two sections.

\subsection{RGB vs RGBD}
\begin{figure}
	\centering
	\includegraphics[width=.8\linewidth]{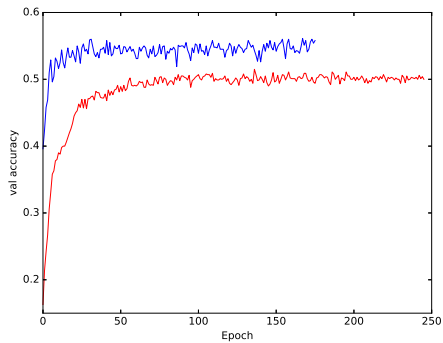}
	\caption{RGB vs RGBD, testing time (RGBD:blue, RGB:red)}
	\label{fig:mixtest}
\end{figure}

%We perform fine tuning for both RGB and RGBD situations.
%So that their performances are approximately optimal.
%Figure~\ref{fig:mixtrain} Compares training accuracy  through time. 
Figure~\ref{fig:mixtest} Compares validation accuracy comparison through time. 
%\begin{figure}
%	\centering
%	\includegraphics[width=\linewidth]{../presentation/together_train.eps}
%	\caption{RGB vs RGBD, training time (RGBD:blue, RGB:red)}
%	\label{fig:mixtrain}
%\end{figure}

We get {\bf56\%} and {\bf52\%} validation accuracy with RGBD and RGB dataset respectively.
This can be seen as a sign that depth map brings extra knowledge learned by deep convolutional neural field to our classification task.

Also notice that, although RGBD dataset have more inputs and neurons,
it has a much higher converge rate than RGB dataset. 
It can be interpreted as a better feature representation brought by depth map.
Estimated depth map works on shallow networks, 
however it remains a question whether it works on deep networks like ResNet \cite{he2015deep}.

\subsection{ResNet Experimentation}
Our previous experiments using a 2-layer feed-forward neural network yielded a performance increase of 4\%,
 when comparing a NN trained on the RGB dataset to the NN trained on the RGBD dataset.
These results did't satisfy us, 
as a simple feed-forward neural network may show how good the features are presented to it, 
but not the optimal tool used for image classification.
We want to see whether the estimated depth map truely bring in some new knowledges of depth, which can't be obtained just using RGB images. 
So we decided to test the performance of ResNet~\cite{he2015deep} over the CIFAR-10 RGB dataset and compare it to the performance achieved over our CIFAR-10 RGBD dataset. Similar with 2-layer networks, only number of channels of input layer is changed from 3 to 4. The increased computational complexity could be ignored, since later layers have much more channels.

Shown in Table~\ref{tab:cnn}, 
ResNet-20 improves \textbf{0.45\%} with estimated depth map.
ResNet-56 achieved error of {\bf6.45\%} using the RGBD dataset, 
which is competitive with ResNet-110 using RGB dataset.
This performance gain of {\bf0.53\%} in accuracy between RGBD and RGB using the CNN could not be ignored on CIFAR-10.
%However, it drops ResNet-110 0.61\%.
%This is counter-intuitive, transfer learning might not always improve the performance.
We conclude that training on RGBD dataset would also result in better performance for deep networks like ResNet.
 %ResNet-110 is really deep enough,
Depth estimation feature may be hard to be generated by deep network itself.
%Additional depth estimation feature may lead to over-fitting, which is similar to curse of dimensionality in machine learning \cite{bellman2015adaptive}.
\begin{table}
	\begin{center}
		\begin{tabular}{|l|l|l|}
			\hline
			Network&RGB & RGBD \\
			\hline
			two-layers&48&\textbf{44}\\
			\hline
			ResNet-20&8.75 & \textbf{8.20}\\
			\hline
			ResNet-56&6.97&\textbf{6.44}\\
			\hline
			%ResNet-110&\textbf{6.43}& 7.04\\			
			%\hline
		\end{tabular}
	\end{center}
	\caption{Performance comparisons of ResNet with or without depth estimation aided on CIFAR-10 (\textit{Smaller is better}).}
	\label{tab:cnn}
\end{table}

\section{Conclusion}
%We successfully reimplemented the state-of-the-art depth-estimation model using
We created a RGBD image dataset for CIFAR-10.
%and investigate its quality using our metric.
We define a transfer learning accuracy metric for depth prediction problem.
On RGBD CIFAR-10, we show that depth channel has a better feature representation.
%However, counter-intuitively 
Training on RGBD images could improve image classification on both shallow and deep networks.

{\small
\bibliographystyle{ieee}
\bibliography{depth}
}

\end{document}